\documentclass[11pt,a4paper]{article}
\usepackage[hyperref]{acl2017}
\usepackage{times}
\usepackage{latexsym,graphicx}

\usepackage{url}

\aclfinalcopy 

\title{``Liar, Liar Pants on Fire'':\\ 
A New Benchmark Dataset for Fake News Detection}

\author{William Yang Wang \\
  Department of Computer Science \\
  University of California, Santa Barbara \\
  Santa Barbara, CA 93106 USA \\
  {\tt william@cs.ucsb.edu}}

\date{}

\begin{document}
\maketitle
\begin{abstract}
Automatic fake news detection is a challenging problem in deception detection, and it has tremendous real-world political and social impacts. However, statistical approaches to combating fake news has been dramatically limited by the lack of labeled benchmark datasets. In this paper, we present \textsc{liar}: a new, publicly available dataset for fake news detection. We collected a decade-long, 12.8K manually labeled short statements in various contexts from \textsc{PolitiFact.com}, which provides detailed analysis report and links to source documents for each case. This dataset can be used for fact-checking research as well. Notably, this new dataset is an order of magnitude larger than previously largest public fake news datasets of similar type. Empirically, we investigate automatic fake news detection based on surface-level linguistic patterns. We have designed a novel, hybrid convolutional neural network to integrate meta-data with text. We show that this hybrid approach can improve a text-only deep learning model.  

\end{abstract}
\section{Introduction}

In this past election cycle for the 45th President of the United States, the world has witnessed a growing epidemic of fake news. The plague of fake news not only poses serious threats to the integrity of journalism, but has also created turmoils in the political world. The worst real-world impact is that fake news seems to create real-life fears: last year, a man carried an AR-15 rifle and walked in a Washington DC Pizzeria, because he recently read online that ``this pizzeria was harboring young children as sex slaves as part of a child-abuse ring led by Hillary Clinton''\footnote{http://www.nytimes.com/2016/12/05/business/media/comet-ping-pong-pizza-shooting-fake-news-consequences.html}. The man was later arrested by police, and he was charged for firing an assault rifle in the restaurant~\cite{nyt2016}.

The broadly-related problem of deception detection~\cite{Mihalcea:2009:LDE:1667583.1667679} is not new to the natural language processing community. A relatively early study by Ott et al.~\shortcite{ott2011finding}
focuses on detecting deceptive review opinions in sentiment analysis, using a crowdsourcing approach to create training data for the positive class, and then combine with truthful opinions from TripAdvisor. Recent studies have also proposed stylometric~\cite{feng2012syntactic}, semi-supervised learning~\cite{haideceptive}, 
and linguistic approaches~\cite{perez2015experiments} to detect deceptive text on crowdsourced datasets.
Even though crowdsourcing is an important approach to create labeled training data, there is a mismatch between training and testing. When testing on real-world review datasets, the results could be suboptimal since the positive training data was created in a completely different, simulated platform. 

The problem of fake news detection is more challenging than detecting deceptive reviews,
since the political language on TV interviews, posts on Facebook and Twitters are mostly short statements. However, the lack of manually labeled fake news dataset is still a bottleneck for advancing computational-intensive, broad-coverage models in this direction. Vlachos and Riedel~\shortcite{vlachos2014fact} are the first to release a public fake news detection and fact-checking dataset, but it only includes 221 statements, which does not permit machine learning based assessments.

To address these issues, we introduce the \textsc{liar} dataset, which includes 12,836 short statements labeled for truthfulness, subject, context/venue, speaker, state, party, and prior history. With such volume and a time span of a decade, \textsc{liar} is an order of magnitude larger than the currently available resources~\cite{vlachos2014fact,ferreira2016emergent} of similiar type. Additionally, in contrast to crowdsourced datasets, the instances in \textsc{liar} are collected in a grounded, more natural context, such as political debate, TV ads, Facebook posts, tweets, interview, news release, etc. In each case, the labeler provides a lengthy analysis report to ground each judgment, and the links to all supporting documents are also provided. 

Empirically, we have evaluated several popular learning based methods on this dataset. The baselines include logistic regression, support vector machines, long short-term memory networks~\cite{hochreiter1997long}, and a convolutional neural network model~\cite{kim:2014:EMNLP2014}. We further introduce a neural network architecture to integrate text and meta-data.
Our experiment suggests that this approach improves the performance of a strong text-only convolutional neural networks baseline. 

\begin{figure}
\centering
{\fontfamily{put}\selectfont
\fbox{\parbox[t][9in]{2.7in}{\textbf{Statement}: \emph{``The last quarter, it was just announced, our gross domestic product was below zero. Who ever heard of this? Its never below zero.''} \\  
\textbf{Speaker}: Donald Trump\\ 
\textbf{Context}: presidential announcement speech\\
\textbf{Label}: Pants on Fire\\
\textbf{Justification}:
According to Bureau of Economic Analysis and National Bureau of Economic Research, the growth in the gross domestic product has been below zero 42 times over 68 years. That’s a lot more than ``never.'' We rate his claim Pants on Fire!
\vspace{1.5ex}
\hrule
\vspace{1.5ex}
\textbf{Statement}: \emph{``Newly Elected Republican Senators Sign Pledge to Eliminate Food Stamp Program in 2015.''}\\
\textbf{Speaker}: Facebook posts\\ 
\textbf{Context}: social media posting\\
\textbf{Label}: Pants on Fire\\
\textbf{Justification}: More than 115,000 social media users passed along a story headlined, ``Newly Elected Republican Senators Sign Pledge to Eliminate Food Stamp Program in 2015.'' But they failed to do due diligence and were snookered, since the story came from a publication that bills itself (quietly) as a ``satirical, parody website.'' We rate the claim Pants on Fire.
\vspace{1.5ex}
\hrule
\vspace{1.5ex}

\textbf{Statement}: \emph{``Under the health care law, everybody will have lower rates, better quality care and better access.''}\\
\textbf{Speaker}:  Nancy Pelosi\\ 
\textbf{Context}: on 'Meet the Press'\\
\textbf{Label}: False\\
\textbf{Justification}: 
Even the study that Pelosi's staff cited as the source of that statement suggested that some people would pay more for health insurance. Analysis at the state level found the same thing. The general understanding of the word ``everybody'' is every person. The predictions don’t back that up. We rule this statement False.
}}}
\caption{Some random excerpts from the \textsc{liar} dataset.}
\label{fig:example}
\end{figure}

\section{\textsc{liar}: a New Benchmark Dataset}
The major resources for deceptive detection of reviews are crowdsourced datasets~\cite{ott2011finding,perez2015experiments}. They are very useful datasets to study deception detection, but the positive training data are collected from a simulated environment. More importantly, these datasets are not suitable for fake statements detection, since the fake news on TVs and social media are much shorter than customer reviews. 

Vlachos and Riedel~\shortcite{vlachos2014fact} are the first to construct fake news and fact-checking datasets. They obtained 221 statements from \textsc{Channel 4}\footnote{http://blogs.channel4.com/factcheck/} and \textsc{PolitiFact.com}\footnote{http://www.politifact.com/}, a Pulitzer Prize-winning website. In particular, PolitiFact covers a wide-range of political topics, and
they provide detailed judgments with fine-grained labels. Recently,
Ferreira and Vlachos~\shortcite{ferreira2016emergent} have released the Emergent dataset, which includes 300 labeled rumors from PolitiFact. However, with less than a thousand samples, it is impractical to use these datasets as benchmarks for developing and evaluating machine learning algorithms for fake news detection. Therefore, it is of crucial significance to introduce a larger dataset to facilitate the development of computational approaches to fake news detection and automatic fact-checking.

\begin{table}[t]
\begin{center}
\begin{tabular}{lr}

\hline
Dataset Statistics &\\
\hline
Training set size & 10,269\\
Validation set size & 1,284\\
Testing set size & 1,283\\
Avg. statement length (tokens) & 17.9 \\
\hline
Top-3 Speaker Affiliations & \\
Democrats & 4,150 \\
Republicans & 5,687 \\
None (e.g., FB posts) & 2,185\\
\hline
\end{tabular}
\caption{The \textsc{liar} dataset statistics.}
\label{tab:stats}
\end{center}
\end{table}

{\sloppypar
We show some random snippets from our dataset in Figure~\ref{fig:example}. The \textsc{liar} dataset\footnote{\url{https://www.cs.ucsb.edu/~william/data/liar\_dataset.zip}} includes 12.8K human labeled short statements from
~\textsc{PolitiFact.com}'s API\footnote{\url{http://static.politifact.com/api/v2apidoc.html}}, and each statement is evaluated by a \textsc{PolitiFact.com} editor for its truthfulness. After initial analysis, we found duplicate labels, and merged the full-flop, half-flip, no-flip labels into false, half-true, true labels respectively. We consider six fine-grained labels for the truthfulness ratings: \emph{pants-fire, false, barely-true, half-true, mostly-true, and true}. 
The distribution of labels in the \textsc{liar} dataset is relatively well-balanced: except for 1,050 pants-fire cases, the instances for all other labels range from 2,063 to 2,638. We randomly sampled 200 instances to examine the accompanied lengthy analysis reports and rulings. 
Not that fact-checking is not a classic labeling task in NLP. The verdict requires extensive training in journalism for finding relevant evidence. Therefore, for second-stage verifications, we went through a randomly sampled subset of the analysis reports to check if we agreed with the reporters' analysis. The agreement rate measured by Cohen’s kappa was 0.82. We show the corpus statistics in Table~\ref{tab:stats}. The statement dates are primarily from 2007-2016.} 

The speakers in the \textsc{liar} dataset include a mix of democrats and republicans,
as well as a significant amount of posts from online social media. We include a rich set of meta-data for each speaker---in addition to party affiliations, current job, home state, and credit history are also provided. In particular, the credit history includes the historical counts of inaccurate statements for each speaker. For example, Mitt Romney has a credit history vector $h=\{19,32,34,58,33\}$,
which corresponds to his counts of ``pants on fire'', ``false'', ``barely true'',  ``half true'', ``mostly true'' for historical statements. Since this vector also includes the count for the current statement, it is important to subtract the current label from the credit history when using this meta data vector in prediction experiments. 

These statements are sampled from various of contexts/venues, and the top categories include \emph{news releases, TV/radio interviews, campaign speeches, TV ads, tweets, debates, Facebook posts, etc}. To ensure a broad coverage of the topics, there is also a diverse set of subjects discussed by the speakers. The top-10 most discussed subjects in the dataset are \emph{economy, health-care, taxes, federal-budget, education, jobs, state-budget, candidates-biography, elections, and immigration}.

\begin{figure}[t]
\centering
\includegraphics[width=1\linewidth]{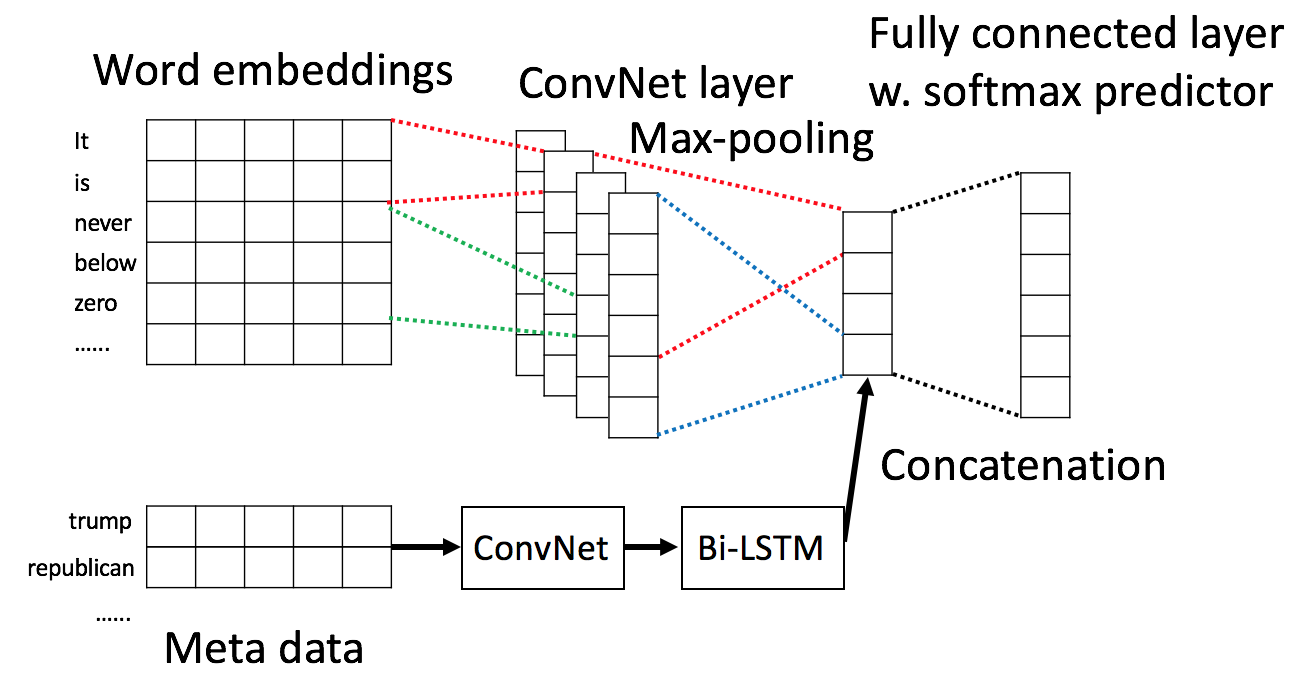}
\caption{The proposed hybrid Convolutional Neural Networks framework for integrating text and meta-data.}
\label{fig:cnn}
\end{figure}

\section{Automatic Fake News Detection}
One of the most obvious applications of our dataset is to facilitate the development of machine learning models for automatic fake news detection. In this task, we frame this as a 6-way multiclass text classification problem. And the research questions are: 
\begin{itemize}
\item Based on surface-level linguistic realizations only, how well can machine learning algorithms classify a short statement into a fine-grained category of fakeness?
\item Can we design a deep neural network architecture to integrate speaker related meta-data with text to enhance the performance of fake news detection?
\end{itemize}

Since convolutional neural networks architectures (CNNs)~\cite{collobert2011natural,kim:2014:EMNLP2014,zhang2015character} have obtained the state-of-the-art results on many text classification datasets, we build our neural networks model based on a recently proposed CNN model~\cite{kim:2014:EMNLP2014}. Figure~\ref{fig:cnn} shows the overview of our hybrid convolutional neural network for integrating text and meta-data. 

We randomly initialize a matrix of embedding vectors to encode the metadata embeddings. We use a convolutional layer to capture the dependency between the meta-data vector(s). Then, a standard max-pooling operation is performed on the latent space, followed by a bi-directional LSTM layer.
We then concatenate the max-pooled text representations with the meta-data representation from the bi-directional LSTM, and feed them to
fully connected layer with a softmax activation function to generate the final prediction.
\section{\textsc{liar}: Benchmark Evaluation}
In this section, we first describe the experimental setup, and the baselines. Then, we present the empirical results and compare various models.
\subsection{Experimental Settings}
We used five baselines: a majority baseline, a regularized logistic regression classifier (LR), a support vector machine classifier (SVM)~\cite{crammer2001algorithmic}, a bi-directional long short-term memory networks model (Bi-LSTMs)~\cite{hochreiter1997long,graves2005framewise}, and a convolutional neural network model (CNNs)~\cite{kim:2014:EMNLP2014}. For LR and SVM, we used the \textsc{LibShortText} toolkit\footnote{https://www.csie.ntu.edu.tw/\~{}cjlin/libshorttext/}, which was shown to provide very strong performances on short text classification problems~\cite{Wang:2015:EMNLP}. For Bi-LSTMs and CNNs, we used TensorFlow for the implementation. We used pre-trained 300-dimensional word2vec embeddings from Google News~\cite{mikolov2013efficient} to warm-start the text embeddings. We strictly tuned all the hyperparameters on the validation dataset. The best filter sizes for the CNN model was (2,3,4). In all cases, each size has 128 filters. The dropout keep probabilities was optimized to 0.8, while no $L_2$ penalty was imposed. The batch size for stochastic gradient descent optimization was set to 64, and the learning process involves 10 passes over the training data for text model. For the hybrid model, we use 3 and 8 as filter sizes, and the number of filters was set to 10. We considered 0.5 and 0.8 as dropout probabilities. The hybrid model requires 5 training epochs. 

We used grid search to tune the hyperparameters for LR and SVM models. We chose accuracy as the evaluation metric, since we found that the accuracy results from various models were equivalent to f-measures on this balanced dataset.
\begin{table}[t]
\begin{center}
\begin{tabular}{lll}

\hline
Models & Valid. & Test\\
\hline
Majority & 0.204 & 0.208\\
SVMs & 0.258 & 0.255\\
Logistic Regress0ion & 0.257 & 0.247\\
Bi-LSTMs & 0.223 & 0.233\\
CNNs & 0.260 & 0.270\\
\hline
Hybrid CNNs& &\\
Text + Subject & 0.263 & 0.235\\
Text + Speaker & \textbf{0.277} & 0.248\\
Text + Job & 0.270 & 0.258\\
Text + State & 0.246 & 0.256\\
Text + Party & 0.259 & 0.248\\
Text + Context & 0.251 & 0.243\\
Text + History & 0.246 & 0.241\\
Text + All & 0.247 & \textbf{0.274}\\
\hline
\end{tabular}
\caption{The evaluation results on the \textsc{liar} dataset. The top section: text-only models. The bottom: text + meta-data hybrid models.}
\label{tab:eval}
\end{center}
\end{table}

\subsection{Results}
We outline our empirical results in Table~\ref{tab:eval}.
First, we compare various models using text features only.
We see that the majority baseline on this dataset gives about 0.204 and 0.208 accuracy on the validation and test sets respectively. 
Standard text classifier such as SVMs and LR models obtained significant improvements. Due to overfitting, the Bi-LSTMs did not perform well. The CNNs outperformed all models, resulting in an accuracy of 0.270 on the heldout test set. We compare the predictions from the CNN model with SVMs via a two-tailed paired t-test, and CNN was significantly better ($p<.0001$).
When considering all meta-data and text, the model achieved the best result on the test data.

\section{Conclusion}
We introduced \textsc{liar}, a new dataset for automatic fake news detection. Compared to prior datasets, \textsc{liar} is an order of a magnitude larger,
which enables the development of statistical and computational approaches to fake news detection. \textsc{liar}'s authentic, real-world short statements from various contexts with diverse speakers also make the research on developing broad-coverage fake news detector possible. We show that when combining meta-data with text, significant improvements can be achieved for fine-grained fake news detection. Given the detailed analysis report and links to source documents in this dataset, it is also possible to explore the task of automatic fact-checking over knowledge base in the future. Our corpus can also be used for stance classification, argument mining, topic modeling, rumor detection, and political NLP research. 
\bibliography{all}
\bibliographystyle{acl_natbib}

\appendix

\end{document}